# Relational Bayesian Networks

Manfred Jaeger*
Computer Science Department,
Stanford University, Stanford CA 94305
jaeger@robotics.stanford.edu

## Abstract

A new method is developed to represent probabilistic relations on multiple random events. Where previously knowledge bases containing probabilistic rules were used for this purpose, here a probability distribution over the relations is directly represented by a Bayesian network. By using a powerful way of specifying conditional probability distributions in these networks, the resulting formalism is more expressive than the previous ones. Particularly, it provides for constraints on equalities of events, and it allows to define complex, nested combination functions.

## 1 INTRODUCTION

In a standard Bayesian network, nodes are labeled with random variables (r.v.s) $X$ that take values in some finite set $\{x_1, \ldots, x_n\}$. A network with r.v.s *(earth)quake, burglary*, and *alarm*, each with possible values $\{true, false\}$, for instance, then defines a joint probability distribution for these r.v.s.

Evidence, $E$, is a set of instantiations of some of the r.v.s. A query asks for the probability of a specific value $x$ of some r.v. $X$, given the instantiations in the evidence. The answer to this query is the conditional probability $P(X = x \mid E)$ in the distribution $P$ defined by the network.

The implicit underlying assumption we here make is that the value assignments in the evidence and the query instantiate the *attributes* of one single random event, or object, that has been sampled (observed) according to the distribution of the network. If, for instance, $E = \{quake = true, alarm = true\}$, then both instantiations are assumed to refer to one single observed state of the world $\omega$, and not the facts that there was an earthquake in 1906, and the alarm bell is ringing right now.

* On leave from: Max-Planck-Institut für Informatik, Im Stadtwald, D-66123 Saarbrücken, Germany

In case we indeed have evidence about several observed events, e.g. $quake(\omega_1) = true$, $alarm(\omega_1) = true$, $burglary(\omega_2) = false$, then, for the purpose of answering a query $X(\omega) = x$ about one of these events, all evidence about other events can be ignored, and only $P(X(\omega) = x \mid E(\omega))$ needs to be computed. For each of these computations the same Bayesian network can be used.

Things become much different when we also want to model *relations* that may hold between two different random events. Suppose, for instance, we also want to say something about the probability that one earthquake was stronger than another. For this we use the binary relation *stronger*, and would like to relate the probability of $stronger(\omega_1, \omega_2)$ to, say, $alarm(\omega_1)$ and $alarm(\omega_2)$. Evidence may now contain instantiations of *stronger* for many different pairs of states: $\{stronger(\omega_1, \omega_2), \ldots, stronger(\omega_1, \omega_n)\}$, and a query may be $alarm(\omega_1)$. In evaluating this query, we no longer can ignore information about the other events $\omega_2, \ldots, \omega_n$. This means, however, that if we do not want to impose an a priori restriction on the number of events we can have evidence for, no single fixed Bayesian network with finite-range r.v.s will be sufficient to evaluate queries for arbitrary evidence sets.

Nevertheless, the probabilistic information that we would like to encode about relations between an arbitrary number of different events may very well be expressible by some finite set of laws, applicable to an arbitrary number of events. One way of expressing such laws, which has been explored in the past ( (Breese 1992),(Poole 1993),(Haddawy 1994)), is to use probabilistic rules such as

$$stronger(u, v) \xleftarrow{0.8} quake(u) \land quake(v)$$
$$\land alarm(u) \land \neg alarm(v). \quad (1)$$

The intended meaning here is: for all states of the world $\omega_1$ and $\omega_2$, given that $quake(\omega_1) \land \ldots \land \neg alarm(\omega_2)$ is true, the probability that $\omega_1$ is stronger than $\omega_2$ is 0.8. A rule-base containing expressions of this form then can be used to construct, for each specific evidence and query, a Bayesian network over binary r.v.s $stronger(\omega_1, \omega_2)$, $stronger(\omega_1, \omega_3)$, $quake(\omega_3)$,..., in which the answer to the query subsequently is computed using standard Bayesian network inference.

In all the above mentioned approaches, quite strong syntactic and/or semantic restrictions are imposed in the formalism that severely limit its expressiveness. Poole (1993) does not allow the general expressiveness of rules like (1), but only combines deterministic rules with the specification of certain unconditional probabilities. Haddawy(1994) allows only rules in which the antecedent does not contain free variables that do not appear in the consequent. As pointed out by Glesner and Koller (1995), this is a severe limitation. For instance, we can then not express by a rule like $aids(x) \xleftarrow{p} contact(x, y)$ that the probability of person $x$ having aids depends on any other person $y$, with whom $x$ had sexual contact. When we do permit an additional free variable $y$ in this manner, it also has to be defined how the probability of the consequent is affected when there exist multiple instantiations of $y$ that make the antecedent true (this question also arises when several rules with the same consequent are permitted in the rule base ). In (Glesner & Koller 1995) and (Ngo, Haddawy & Helwig 1995) therefore a *combination rule* is added to the rule-base, which defines how the conditional probabilities arising from different instantiations, or rules, are to be combined. If the different causal relationships described by the rules are understood to be independent, then the combination rule typically will be noisy-or.

The specification of a single combination rule applied to all sets of instantiations of applicable rules, again, does not permit us to describe certain important distinctions. If, for instance, we have a rule that relates $aids(x)$ to the relation $contact(x, y)$, and another rule that relates $aids(x)$ to the relation $donor(x, y)$, standing for the fact that $x$ has received a blood transfusion from donor $y$, then the probability computed for $aids(a)$, using a simple combination rule, will depend only on the number of instantiations for $contact(a, y)$ and for $donor(a, y)$. Particularly, we are not able to make special provisions for the two rules to be instantiated by the same element $b$, even though the case $contact(a, b) \wedge donor(a, b)$ clearly has to be distinguished from the case $contact(a, b) \wedge donor(a, c)$, or even $contact(a, b) \wedge donor(a, a)$.

In this paper a representation formalism is developed that incorporates constraints on the equality of instantiating elements, and thereby allows us to define different probabilities in situations only distinguished by equalities between instantiating elements.

Furthermore, our representation method will allow us to specify hierarchical, or nested, combination rules. As an illustrations of what this means, consider the unary predicate $cancer(x)$, representing that person $x$ will develop cancer at some time, and the three placed relation $exposed(x, y, z)$, representing that organ $y$ of person $x$ was exposed to radiation at time $z$ (by the taking of an x-ray, intake of radioactively contaminated food, etc). Suppose, now, that for person $x$ we have evidence $E = \{exposed(x, y_i, z_j) \mid i = 1, \ldots, k; j = 1, \ldots, l\}$, where $y_i = y_{i'}$ for some $i, i'$, and $z_j = z_{j'}$ for some $j, j'$. Assume that for any specific organ $y$, multiple exposures of $y$ to radiation have a cumulative effect on the risk of developing cancer of $y$, so that noisy-or is not the adequate rule to model the combined effect of instances $exposed(x, y, z_j)$ on the probability of developing cancer of $y$. On the other hand, developing cancer at any of the various organs $y$ can be viewed as independent causes for developing cancer at all. Thus, a single rule of the form $cancer(x) \xleftarrow{p} exposed(x, y, z)$ together with a "flat" combination rule is not sufficient to model the true probabilistic relationships. Instead, we need to use one rule to first combine for every fixed $y$ the instances given by different $z$, and then use another rule (here noisy-or) to combine the effect of the different $y$'s.

To permit constraints on the equality of instantiating elements, and to allow for hierarchical definitions of combination functions, in this paper we depart from the method of representing our information in a knowledge base containing different types of rules. Instead, we here use Bayesian networks with a node for every relation symbol $r$ of some vocabulary $S$, which is seen as a r.v. whose values are possible interpretations of $r$ in some specific domain $D$. The state space of these *relational Bayesian networks* therefore can be identified with the set of all $S$-structures over $D$, and its semantics is a probability distribution over $S$-structures, as were used by Halpern(1990) to interpret first-order probabilistic logic. Halpern and Koller(1996) have used Markov networks labeled with relation symbols for representing conditional independencies in probability distributions over $S$-structures. This can be seen as a qualitative analog to the quantitative relational Bayesian networks described here.

## 2   THE BASIC FRAMEWORK

In medical example domains it is often natural to make the domain closure assumption, i.e. to assume that the domain under consideration consists just of those objects mentioned in the knowledge base. The following example highlights a different kind of situation, where a definite domain of objects is given over which the free variables are to range, yet there is no evidence about most of these objects.

**Example 2.1** Robot TBayes0.1 moves in an environment consisting of $n$ distinct locations. TBayes0.1 can make direct moves from any location $x$ to any location $y$ unless the (directed) path $x \rightarrow y$ is *blocked*. This happens to be the case with probability $p_0$ for all $x \neq y$. At each time step TBayes0.1, as well as a certain number of other robots op-erating in this domain, make one move along an unblocked path $x \to y, x \neq y$. TBayes0.1 just has completed the task it was assigned to do, and is now in search of new instructions. It can receive these instructions, either by reaching a *terminal*-location from where a central task assigning computer can be accessed, or by meeting another robot that will assign TBayes0.1 a subtask of its own job. Unfortunately, TBayes0.1 only has the vaguest idea of where the terminal locations are, or where the other robots are headed. The best model of its environment that it can come up with, is that every location $x$ is a terminal location with probability $p_1$, and that any unblocked path $x \to y$ is likely to be taken by at least one robot at any given time step with probability $p_2$. In order to plan its next move, TBayes0.1 tries to evaluate for every location $x$ the probability that going to $x$ leads to *success*, defined as either getting instructions at $x$ directly, or being able to access a terminal location in one more move from $x$. Hence, the probability of $s(uccess)(x)$ is 1 if $t(erminal)(x)$ is true, or if $t(z)$ and $\neg b(locked)(x, z)$ holds for some $z$. Otherwise, there still is a chance of $s(x)$ being true, determined by the number of incoming paths $z \to x$, each of which is likely to be taken by another robot with probability $p_2$. Assuming a fairly large number of other robots, the event that $z \to x$ is taken by some robot can be viewed as independent from $z' \to x$ being taken by a robot, so that the overall probability that another robot will reach location $x$ is given by $1 - (1 - p_2)^k$, where $k = |\{z \mid z \neq x, \neg b(z, x)\}|$, i.e. by combining the individual probabilities via noisy-or.

The foregoing example gives an informal description of how the probability of $s(x)$ is evaluated, given the predicates $b$ and $t$. Also, the probabilities for $b$ and $t$ are given. Piecing all this information together (and assuming independence whenever no dependence has been mentioned explicitly), we obtain for every finite domain $D$ of locations a probability distribution $P$ for the $\{b, t, s\}$-structures over $D$.

Our aim now is to represent this class of probability distributions in compact form as a Bayesian network with nodes $b, t$, and $s$. Given the description of the dependencies in the example, it is clear that this network should have two edges: one leading from $b$ to $s$, and one leading from $t$ to $s$.

The more interesting problem is how to specify the conditional probability of the possible values of each node (i.e. the possible interpretations of the symbol at that node), given the values of its parent nodes. For the two parentless nodes in our example this is accomplished very easily: for a given domain $D$, and for all locations $x, y \in D$ we have

$$P(b(x,y)) = \begin{cases} p_0 & \text{if } x \neq y \\ 0 & \text{if } x = y \end{cases} \quad (2)$$

$$P(t(x)) = p_1. \quad (3)$$

Here $P(b(x, y))$ stands for the probability that $(x, y)$ belongs to the interpretation of $b$. Similarly for $P(t(x))$. Since $b(x, y)$ and $b(x', y')$ for $(x, y) \neq (x', y')$, respectively $t(x)$ and $t(x')$ for $x \neq x'$, were assumed to be mutually independent, this defines a probability distribution over the possible interpretations in $D$ of the two predicates. For example, the probability that $I \subseteq D \times D$ is the interpretation of $b$ is 0 if $(x, x) \in I$ for some $x \in D$, and $p_0^{|I|}(1 - p_0)^{n(n-1)-|I|}$ else.

Next, we have to define the probability of interpretations of $s$. Given interpretations of $b$ and $t$, the events $s(x)$ and $s(x')$ are independent for $x \neq x'$. Also, example 2.1 contains a hight level description of how the probability of $s(x)$ is to be computed. Our aim now is to formalize this computation rule in such a manner, that $P(s(x))$ can be computed by evaluating a single functional expression, in the same manner as $P(b(x, y))$ and $P(t(x))$ are given by (2) and (3).

Since $P(s(x))$ depends on the interpretations of $b$ and $t$, we begin with functional expressions that access these interpretations. This is done by using indicator functions $1_{I(b)}(x, y)$ and $1_{I(t)}(x)$. $1_{I(b)}(x, y)$, for example, evaluates to 1 if $(x,y)$ is in the given interpretation $I(b)$ of $b$, and to 0 otherwise. Though the function $1_{I(b)}(x, y)$ has to be distinguished from the logical expression $b(x, y)$, for the benefit of greater readability, in the sequel the simpler notation will be used for both. Thus, $b(x, y)$ stands for the function $1_{I(b)}(x, y)$ whenever it appears within a functional expression.

In order to find a suitable functional expression $F_s(x)$ for $P(s(x))$, assume first that $t(x)$ is true. Since $t(x)$ implies $s(x)$, in this case we need to obtain $F_s(x) = 1$. In the case $\neg t(x)$, the probability of $s(x)$ is computed by considering all locations $z \neq x$ for which either $\neg b(x, z)$ or $\neg b(z, x)$. Any such $z$ that satisfies $\neg b(x, z) \wedge t(z)$ again makes $s(x)$ true with probability 1. If only $\neg b(z, x)$ holds, then the location $z$ merely "contributes" a probability $p_2$ to $P(s(x))$. Thus, for any $z$, the contribution of $z$ to $P(s(x))$ is given by $max\{t(z)(1 - b(x, z)), p_2(1 - b(z, x))\}$. Combining all the relevant $z$ via noisy-or, we obtain the formula

$$F_s(x) = n\text{-}o\{max\{t(z)(1 - b(x, z)), p_2(1 - b(z, x))\} \mid z; z \neq x\} \quad (4)$$

for $x$ with $\neg t(x)$.

Abbreviating the functional expression on the right-hand side of (4) by $H(x)$, we can finally put the two cases $t(x)$ and $\neg t(x)$ together, defining

$$F_s(x) = t(x) + (1 - t(x))H(x). \quad (5)$$

We now give a general definition of a representation language for forming functional expressions in the style of (5). We begin by describing the general class of *combination functions*, instances of which are the functions $n\text{-}o$ and $max$ used above.

**Definition 2.2** A combination function is any function that maps every finite multiset (i.e. a set possibly containing multiple copies of the same element) with elements from [0,1] into [0,1].

Except *n-o* and *max*, examples of combination functions are *min*, the arithmetic mean of the arguments, etc. Each combination function must include a sensible definition for its result on the empty set. For example, we here use the conventions $n\text{-}o\,\emptyset = max\,\emptyset = 0$, $min\,\emptyset = 1$.

In the following, we use bold type to denote tuples of variables: $\boldsymbol{x} = (x_1, \ldots, x_n)$ for some $n$. The number of elements in tuple $\boldsymbol{x}$ is denoted by $|\boldsymbol{x}|$. An *equality constraint* $c(\boldsymbol{x})$ for $\boldsymbol{x}$ is a quantifier free formula over the empty vocabulary, i.e., a formula only containing atomic subformulas of the form $x_i = x_j$.

**Definition 2.3** The class of *probability formulas* over the relational vocabulary $S$ is inductively defined as follows.

(i) (Constants) Each rational number $q \in [0, 1]$ is a probability formula.

(ii) (Indicator functions) For every $n$-ary symbol $r \in S$, and every $n$-tuple $\boldsymbol{x}$ of variables, $r(\boldsymbol{x})$ is a probability formula.

(iii) (Convex combinations) When $F_1, F_2, F_3$ are probability formulas, then so is $F_1 F_2 + (1 - F_1)F_3$.

(iv) (Combination functions) When $F_1, \ldots, F_k$ are probability formulas, *comb* is any combination function, $\boldsymbol{x}, \boldsymbol{z}$ are tuples of variables, and $c(\boldsymbol{x}, \boldsymbol{z})$ is an equality constraint, then $comb\{F_1, \ldots, F_k \mid \boldsymbol{z}; c(\boldsymbol{x}, \boldsymbol{z})\}$ is a probability formula.

Note that special cases of (iii) are multiplication ($F_3 = 0$) and "inversion" ($F_2 = 0$, $F_3 = 1$). The set of free variables of a probability formula is defined in the canonical way. The free variables of $comb\{\ldots\}$ are the union of the free variables of the $F_i$, minus the variables in $\boldsymbol{z}$.

A probability formula $F$ over $S$ in the free variables $\boldsymbol{x} = (x_1, \ldots, x_n)$ defines for every $S$-structure $\mathscr{D}$ over a domain $D$ a mapping $D^n \mapsto [0, 1]$. The value $F(\boldsymbol{d})$ for $\boldsymbol{d} \in D^n$ is defined inductively over the structure of $F$. We here give the details only for case (iv).

Let $F(\boldsymbol{x})$ be of the form $comb\{F_1(\boldsymbol{x}, \boldsymbol{z}), \ldots, F_k(\boldsymbol{x}, \boldsymbol{z}) \mid \boldsymbol{z}; c(\boldsymbol{x}, \boldsymbol{z})\}$ (where not necessarily all the variables in $\boldsymbol{x}$ and $\boldsymbol{z}$ actually appear in all the $F_i$ and in $c$). In order to define $F(\boldsymbol{d})$, we must specify the multiset represented by

$$\{F_1(\boldsymbol{d}, \boldsymbol{z}), \ldots, F_k(\boldsymbol{d}, \boldsymbol{z}) \mid \boldsymbol{z}; c(\boldsymbol{d}, \boldsymbol{z})\}. \quad (6)$$

Let $E \subseteq D^{|\boldsymbol{z}|}$ be the set $\{\boldsymbol{d}' \mid c(\boldsymbol{d}, \boldsymbol{d}')\}$. For each $\boldsymbol{d}' \in E$ and each $i \in \{1, \ldots, k\}$, by induction hypothesis, $F_i(\boldsymbol{d}, \boldsymbol{d}') \in [0, 1]$. The multiset represented by (6) now is defined as containing as many copies of $p \in [0, 1]$ as there are representations $p = F_i(\boldsymbol{d}, \boldsymbol{d}')$ with different $i$ or $\boldsymbol{d}'$. Note that $F_i(\boldsymbol{d}, \boldsymbol{d}')$ and $F_i(\boldsymbol{d}, \boldsymbol{d}'')$ count as different representations even in the case that the variables for which $\boldsymbol{d}'$ and $\boldsymbol{d}''$ substitute different elements do not actually appear in $F_i$. The multiset $\{r(\boldsymbol{d}) \mid z; z = z\}$, for instance, contains as many copies of the indicator $r(\boldsymbol{d})$, as there are elements in the domain over which it is evaluated.

For any tautological constraint like $z = z$, in the sequel we simply write $\tau$.

Another borderline case that needs clarification is the case where $\boldsymbol{z}$ is empty. Here our definition degenerates to: if $c(\boldsymbol{d})$ holds, then the multiset $\{F_1(\boldsymbol{d}), \ldots, F_k(\boldsymbol{d}) \mid \emptyset; c(\boldsymbol{d})\}$ contains as many copies of $p \in [0, 1]$ as there are representations $p = F_i(\boldsymbol{d})$; it is empty if $c(\boldsymbol{d})$ does not hold.

By using indicator functions $r(\boldsymbol{x})$, the value of $F(\boldsymbol{d})$ is being defined in terms of the validity in $\mathscr{D}$ of atomic formulas $r(\boldsymbol{d}')$. A natural generalization of probability formulas might therefore be considered, in which not only the truth values of atomic formulas are used, but indicator functions for arbitrary first-order formulas are allowed. As the following lemma shows, this provides no real generalization.

**Lemma 2.4** Let $\phi(\boldsymbol{x})$ be a first-order formula over the relational vocabulary $S$. Then there exists a probability formula $F_\phi(\boldsymbol{x})$ over $S$, using *max* as the only combination function, s.t. for every finite $S$-structure $\mathscr{D}$, and every $\boldsymbol{d} \in D^{|\boldsymbol{x}|}$: $F_\phi(\boldsymbol{d}) = 1$ iff $\phi(\boldsymbol{d})$ holds in $\mathscr{D}$, and $F_\phi(\boldsymbol{d}) = 0$ else.

**Proof:** By induction on the structure of $\phi$. If $\phi \equiv r(\boldsymbol{x})$ for some $r \in S$, then $F_\phi(\boldsymbol{x}) = r(\boldsymbol{x})$. For $\phi \equiv x_1 = x_2$, let $F_\phi(x_1, x_2) = max\{1 \mid \emptyset; x_1 = x_2\}$. Conjunction and negation are handled by multiplication and inversion, respectively, of probability formulas. For $\phi \equiv \exists y \psi(\boldsymbol{x}, y)$ the corresponding probability formula is $F_\phi(\boldsymbol{x}) = max\{F_\psi(\boldsymbol{x}, y) \mid y; \tau\}$. □

**Definition 2.5** A *relational Bayesian network* for the (relational) vocabulary $S$ is given by a directed acyclic graph containing one node for every $r \in S$. The node for an $n$-ary $r \in S$ is labeled with a probability formula $F_r(x_1, \ldots, x_n)$ over the symbols in the parent nodes of $r$, denoted by $Pa(r)$.

The definition for the probability of $b(x, y)$ in (2) does not seem to quite match definition 2.5, because it contains a distinction by cases not accounted for in definition 2.5. However, this distinction by cases can be incorporated into a single probability formula. If, for instance, $c_1(\boldsymbol{x})$ and $c_2(\boldsymbol{x})$ are two mutually exclusive and exhaustive equality constraints, then

$$F(\boldsymbol{x}) := max\{max\{F_1(\boldsymbol{x}) \mid \emptyset; c_1(\boldsymbol{x})\}, \\ max\{F_2(\boldsymbol{x}) \mid \emptyset; c_2(\boldsymbol{x})\} \mid \emptyset; \tau\} \quad (7)$$

evaluates to $F_1(x)$ for $x$ with $c_1(x)$, and to $F_2(x)$ for $x$ with $c_2(x)$.

Let $N$ now be a relational Bayesian network over $S$. Let $r$ be (the label of) a node in $N$ with arity $n$, and let $\mathcal{D}$ be a $Pa(r)$-structure over domain $D$. For every $d \in D^n$, $F_r(d) \in [0, 1]$ then is defined. Thus, for every interpretation $I(r)$ of $r$ in $D^n$ we can define

$$P(I(r)) := \prod_{d \in I(r)} F_r(d) \prod_{d \notin I(r)} (1 - F_r(d)),$$

which gives a probability distribution over interpretations of $r$, given the interpretations of $Pa(r)$. Given a fixed domain $D$, a relational Bayesian network thus defines a joint probability distribution $P$ over the interpretations in $D$ of the symbols in $S$, or, equivalently, a probability measure on $S$-structures over $D$. Hence, semantically, relational Bayesian networks are mappings of finite domains $D$ into probability measures on $S$-structures over $D$.

**Example 2.6** Reconsider the relations *cancer* and *exposed* as described in the introduction. Assume that $\gamma : \mathbf{N} \to [0, 1]$ is the probability distribution that for any fixed organ $y$ gives the probability that $y$ develops cancer after the $n$th exposure to radiation. Let $\Gamma(n) := \sum_{i=0}^{n} \gamma(n)$ be the corresponding distribution function. Then $\Gamma$ can be used to define a combination function $comb_\Gamma$ by letting for a multiset $A$: $comb_\Gamma A := \Gamma(n)$, where $n$ is the number of nonzero elements in $A$ (counting multiplicities). Using $comb_\Gamma$ we obtain the probability formula $comb_\Gamma \{exposed(x, y, z) \mid z; \tau\}$ for the contribution of organ $y$ to the cancer risk of $x$. Combining for all $y$, then

$$Fcancer(x) = n\text{-}o\{comb_\Gamma\{exposed(x, y, z) \mid z; \tau\} \mid y; \tau\}$$

is a probability formula defining the risk of cancer for $x$, given the relation *exposed*.

In the preceding example we have tacitly assumed a multi-sorted domain, so that the variables $x, y, z$ range over different sets "people", "organs", "times", respectively. We here do not introduce an extra formalization for dealing with many sorted domains. It is clear that this can be done easily, but would introduce an extra load of notation.

## 3 INFERENCE

The inference problem we would like to solve is: given a relational Bayesian network $N$ for $S$, a finite domain $D = \{d_1, \ldots, d_n\}$, an evidence set of ground literals $E = \{r_1(d_1), \ldots, r_k(d_k), \neg r_{k+1}(d_{k+1}), \ldots, \neg r_m(d_m)\}$ with $r_i \in S$ (not necessarily distinct), $d_i \subseteq D$ (not necessarily distinct) for $i = 1, \ldots, m$, and a ground atom $r_0(d_0)$ ($r_0 \in S, d_0 \subseteq D$), what is the probability of $r_0(d_0)$ given $r_1(d_1), \ldots, \neg r_m(d_m)$? More precisely: in the probability measure $P$ defined by $N$ on the $S$- structures over

$D$, what is the conditional probability $P(r_0(d_0) \mid E)$ of a structure satisfying $r_0(d_0)$, given that it satisfies $r_1(d_1), \ldots, \neg r_m(d_m)$?

Since for any given finite domain a relational Bayesian network can be seen as an ordinary Bayesian network for variables with finitely many possible values, in principle, any inference algorithm for standard Bayesian networks can be used.

Unfortunately, however, direct application of any such algorithm will be inefficient, because they include a summation over all possible values of a node, and the number of possible values here is exponential in the size of the domain. For this reason, it will often be more efficient to follow the approach used in inference from rule-base encodings of probabilistic knowledge, and to construct for every specific inference task an auxiliary Bayesian network whose nodes are ground atoms in the symbols from $S$, each of which with the two possible values *true* and *false* (cf.(Breese 1992),(Ngo et al. 1995)).

The reason why we here can do the same is that in the query $r_0(d_0)$ we do not ask for the probability of any specific interpretation of $r_0$, but only for the probability of all interpretations containing $d_0$. For the computation of this probability, in turn, it is irrelevant to know the exact interpretations of parent nodes $r'$ of $r$. Instead, we only need to know which of those tuples $d'$ belong to $r'$, whose indicator $r'(d')$ is needed in the computation of $F_{r_0}(d_0)$.

In order to construct such an auxiliary network, we have to compute for some given atom $r(d)$ the list of atoms $r'(d')$ on whose truth value $F_r(d)$ depends. One way of doing this is to just go through a recursive evaluation of $F_r(d)$, and list all the ground atoms encountered in this evaluation. However, rather than doing this, it is useful to compute for every relation symbol $r \in S$, and each parent relation $r'$ of $r$, an explicit description of the tuples $y$, such that $F_r(x)$ depends on $r'(y)$. Such an explicit description can be given in form of a first-order formula $pa_{rr'}(x, y)$ over the empty vocabulary.

To demonstrate the general method for the computation of these formulas, we show how to obtain $pa_{sb}(x, y_1, y_2)$ for $F_s(x)$ as defined in (5). By induction on the structure of $F_s$, we compute formulas $pa_{Gb}(x, y_1, y_2)$ that define for a subformula $G(x)$ of $F_s$ the set of $(y_1, y_2)$ s.t. $G(x)$ depends on $b(y_1, y_2)$. In the end, then, $pa_{sb}(x, y_1, y_2) :\equiv pa_{F_s b}(x, y_1, y_2)$.

The two subformulas $t(x)$ and $(1 - t(x))$ of $F_s$ do not depend on $b$ at all; therefore we can let $pa_{t(x)b}(x, y_1, y_2) \equiv pa_{(1-t(x))b}(x, y_1, y_2) \equiv \epsilon$, where $\epsilon$ is some unsatisfiable formula.

To obtain $pa_{H(x)b}(x, y_1, y_2)$ we begin with the atomic subformulas $b(x, z)$ and $b(z, x)$ of $H(x)$, which yield $pa_{b(x,z)b}(x, z, y_1, y_2) \equiv y_1 = x \land y_2 = z$ and $pa_{b(z,x)b}(x,z,y_1,y_2) \equiv y_1 = z \wedge y_2 = x$ respectively. The remaining atomic subformulas $t(z)$, 1, and $p_2$ appearing within the *max* combination function again only yield the unsatisfiable $\epsilon$. Skipping one trivial step where the formulas for the two arguments of $M(x,z) :\equiv max\{\ldots\}$ are computed, we next obtain the formula

$$pa_{M(x,z)b}(x,z,y_1,y_2) \equiv \\ (y_1 = x \wedge y_2 = z) \vee (y_1 = z \wedge y_2 = x)$$

(after deleting some meaningless $\epsilon$-disjuncts). $H(x) = n\text{-}o\{M(x,z) \mid z; z \neq x\}$ depends on all $b(y_1, y_2)$ for which there exist some $z \neq x$ s.t. $pa_{M(x,z)b}(x,z,y_1,y_2)$. Hence,

$$pa_{H(x)b}(x,y_1,y_2) \equiv \\ \exists z((y_1 = x \wedge y_2 = z) \vee (y_1 = z \wedge y_2 = x)), \quad (8)$$

which is already the same as $pa_{F_s(x)b}(x,y_1,y_2)$. Finally, we can simplify (8), and obtain

$$pa_{sb}(x,y_1,y_2) \equiv \\ (y_1 = x \wedge y_2 \neq x) \vee (y_1 \neq x \wedge y_2 = x). \quad (9)$$

In general, the formulas $pa_{rr'}(x,y)$ are existential $\emptyset$-formulas. It is not always possible to completely eliminate the existential quantifiers as in the preceding example. However, it is always possible to transform $pa_{rr'}(x,y)$ into a formula so that quantifiers only appear in subformulas of the form $\exists^{\geq n} x x = x$, postulating the existence of at least $n$ elements. This means that for every formula $pa_{rr'}(x,y)$, and tuples $d, d' \subseteq D$, it can be checked in time independent of the size of $D$ whether $pa_{rr'}(d, d')$ holds.

The formula $pa_{rr'}(x,y)$ enables us to find for every tuple $d$ the parents $r'(d')$ of $r(d)$ in the auxiliary network. Moreover, we can take this one step further: suppose that in the original network $N$ there is a path of length two from node $r''$ via a node $r'$ to $r$. Then, in the auxiliary network, there is a path of length two from a node $r''(d'')$ via a node $r'(d')$ to $r(d)$ iff the formula

$$pa_{r'' \to r' \to r}(x,y) :\equiv \exists z(pa_{rr'}(x,z) \wedge pa_{r'r''}(z,y)). \quad (10)$$

is satisfied for $x = d$, and $y = d''$. Taking the disjunction of all formulas of the form (10) for all paths in $N$ leading from $r''$ to $r$ then yields a formula $pa^*_{rr''}(x,y)$ defining all predecessors $r''(d'')$ of a node $r(d)$ in the auxiliary network.

Using the $pa_{rr'}$ and $pa^*_{rr''}$, we can for given evidence and query construct the auxiliary network needed to answer the query: we begin with a node $r_0(d_0)$ for the query. For all nodes $r(d)$ added to the network, we add all parents $r'(d')$ of $r(d)$, as defined by $pa_{rr'}$. If $r(d)$ is not instantiated in $E$, using the formulas $pa^*_{r'r}$, we check whether the subgraph rooted at $r(d)$ contains a node instantiated in $E$. If this is the case, we add all successors of $r(d)$ that lie on a path from $r(d)$ to an instantiated node (these are again given by the formulas $pa^*_{r'r}$). Thus, we can construct directly the minimal network needed to answer the query, without first backward chaining from every atom in $E$, and pruning afterwards.

Auxiliary networks as described here still encode finer distinctions in the instantiations of the nodes of $N$ than is actually needed to solve our inference problem. Consider, for example, the case where the domain in example 2.1 consists of ten locations $\{l_1, \ldots, l_{10}\}$, there is no evidence, and the query is $s(l_1)$. According to (9), the auxiliary network will contain nodes $b(l_1, l_i), b(l_i, l_1)$ for all $i = 2, \ldots, 10$. In applying standard inference techniques on this network, we distinguish e.g. the case where $b(l_1, l_2), b(l_2, l_1)$ are true and $b(l_1, l_3), b(l_3, l_1)$ are false from the case where $b(l_1, l_2), b(l_2, l_1)$ are false and $b(l_1, l_3), b(l_3, l_1)$ are true, and all other $b(l_1, l_i), b(l_i, l_1)$ have the same truth value. However, for the given inference problem, this distinction really is unnecessary, because the identity of locations mentioned neither in evidence nor query is immaterial. Future work will therefore be directed towards finding inference techniques for relational Bayesian networks that distinguish instantiations of the relations in the network at a higher level of abstraction than the current auxiliary networks, and thereby reduce the complexity of inference in terms of the size of the underlying domain.

## 4  RECURSIVE NETWORKS

In the distributions defined by relational Bayesian networks of definition 2.5, the events $r(a)$ and $r(a')$ with $a \neq a'$ are conditionally independent, given the interpretation of the parent relations of $r$. This is a rather strong limitation of the expressiveness of these networks. For instance, using these networks, we can not model a variation of example 2.1 in which the predicate *blocked* is symmetric: $b(x, y)$ being independent from $b(y, x)$, $b(x, y) \Leftrightarrow b(y, x)$ can not be enforced.

There are other interesting things that we are not able to model so far. Among them are random functions (the main concern of (Haddawy 1994)), and a recursive temporal dependence of a relation on itself (addressed both in (Ngo et al. 1995) and (Glesner & Koller 1995)). In this section we define a straightforward generalization of relational Bayesian networks that allows us to treat all these issues in a uniform way.

We can identify a recursive dependence of a relation on itself as the general underlying mechanism we have to model. In the case of symmetric relations, this is a dependence of $r(x, y)$ on $r(y, x)$. In the case of a temporal development, this is the dependence of a predicate $r(t, x)$, having a time-variable as its first argument, on $r(t-1, x)$. Functions can be seen as special relations $r(x, y)$, where for every $x$ there exists exactly one $y$, s.t. $r(x, y)$ is true. Thus, for every $x$, $r(x, y)$ depends on all $r(x, y')$ in that exactly one of these atoms must be true.

It is clear that there is no fundamental problem in modeling such recursive dependencies within a Bayesian network framework, as long as the recursive dependency of $r(x)$ on $r(y_1), \ldots r(y_k)$ does not produce any cycles. Most obviously, in the case of a temporal dependency, the use of $r(t-1, x)$ in a definition of the probability of $r(t, x)$ does not pose a problem, as long as a non-recursive definition of the probability of $r(0, x)$ is provided.

To make the recursive dependency of $r(x, y)$ on $r(y, x)$ in a symmetric relation similarly well-founded, we can use a total order $\leq$ on the domain. Then we can generate a random symmetric relation by first defining the probability of $r(x, y)$ with $x \leq y$, and then the (0,1-valued) probability of $r(y, x)$ given $r(x, y)$. Now consider the case of a random function $r(x, y)$ with possible values $y \in \{v_1, \ldots, v_k\}$. Here, too, we can make the interdependence of the different $r(x, y)$ acyclic by using a total order on $\{v_1, \ldots, v_k\}$, and assigning a truth value to $r(x, v_i)$ by taking into account the already defined truth values of $r(x, v_j)$ for all $v_j$ that precede $v_i$ in that order.

From these examples we see that what we essentially need, in order to extend our framework to cover a great variety of interesting specific forms of probability distributions over $S$-structures, are well-founded orderings on tuples of domain elements. These well-founded orderings can be supplied via rigid relations on the domain, i.e. fixed, predetermined relations that are not generated probabilistically. Indeed, one such relation we already have used throughout: the equality relation. It is therefore natural to extend our framework by allowing additional relations that are to be used in the same way as the equality predicate has been employed, namely, in constraints for combination functions. Also, fixed constants will be needed as the possible values of random functions.

For the case of a binary symmetric relation $r(x, y)$, assume, as above, that we are given a total (non-strict) order $\leq$ on the domain. A probability formula that defines a probability distribution concentrated on symmetric relations, and making $r(x_1, x_2)$ true with probability $p$ for all $(x_1, x_2)$, then is

$$F_r(x_1, x_2) = max\{max\{p \mid \emptyset; x_1 \leq x_2\}, \quad (11)$$
$$max\{r(x_2, x_1) \mid \emptyset; \neg x_1 \leq x_2\} \mid \emptyset; \tau\}.$$

As in (7), here a nested $max\{\ldots\}$-function is used in order to model a distinction by cases. The first inner $max$-function evaluates to $p$ if $x_1 \leq x_2$, and to 0 else. The second $max$-function is equal to $r(x_2, x_1)$ if $x_1 > x_2$, and 0 else.

For the temporal example, assume that the domain contains $n+1$ time points $t_0, \ldots, t_n$, and a successor relation $s = \{(t_i, t_{i+1}) \mid 0 \leq i \leq n-1\}$ on the $t_i$'s. Assume that $r(t, x)$ is a relation with a time parameter as the first argument, and that $r(t_0, x)$ shall hold with probability $p_0$ for all $x$, whereas $r(t_{i+1}, x)$ has probability $p_1$ if $r(t_i, x)$ holds, and probability $p_2$ else. In order to define the probability of $r(t, x)$ by a probability formula, the case $t = t_0$ must be distinguished from the case $t = t_i$, $i \geq 1$. For this we use the probability formula $F_0(t) = max\{1 \mid t'; s(t', t)\}$, which evaluates to 0 for $t = t_0$, and to 1 for $t = t_1, \ldots, t_n$. We can now use the formula

$$F_r(t, x) = (1 - F_0(t))p_0 +$$
$$F_0(t)max\{r(t', x)p_1 + (1 - r(t', x))p_2$$
$$\mid t'; s(t', t)\}$$

to define the probability of $r(t, x)$.

Finally, for a functional relation $r(x, y)$, suppose that we are given a domain, together with the interpretations of $n$ constant symbols $v_1, \ldots, v_n$, and a strict total order $<$, s.t. $v_1 < v_2 < \ldots < v_n$. Now consider the probability formula

$$F_r(x, y) = (1 - max\{r(x, z) \mid z; z < y\}) \cdot$$
$$max\{max\{p_1 \mid \emptyset; y = v_1\}, \ldots,$$
$$max\{p_n \mid \emptyset; y = v_n\} \mid \emptyset; \tau\}$$

The first factor in this formula tests whether $r(x, z)$ already is true for some possible value $v_i < y$. If this is the case, then the probability of $r(x, y)$ given by $F_r(x, y)$ is 0. Otherwise, the probability of $r(x, y)$ is $p_i$ iff $y = v_i$. The probability that by this procedure the argument $x$ is assigned the value $v_i$ then is $(1-p_1)(1-p_2)\ldots(1-p_{i-1})p_i$. By a suitable choice of the $p_i$ any probability distribution over the $v_i$ can be generated.

The given examples motivate a generalization of relational Bayesian networks. For this, let $R$ be a vocabulary containing relation and constant symbols, $S$ a relational vocabulary with $R \cap S = \emptyset$. An *R-constraint* $c(x)$ *for* $x$ is a quantifier-free $R$-formula. Define the class of *R-probability formulas over $S$* precisely as in definition 2.3, with "equality constraint" replaced by "$R$-constraint".

**Definition 4.1** Let $R, S$ be as above. A *recursive relational Bayesian network for $S$ with $R$-constraints* is given by a directed acyclic graph containing one node for every $r \in S$. The node for an $n$-ary $r \in S$ is labeled with an $R$-probability formula $F_r(x_1, \ldots, x_n)$ over $Pa(r) \cup \{r\}$.

The semantics of a recursive relational Bayesian network is a bit more complicated than that of relational Bayesian networks. The latter defined a mapping of domains $D$ into probability measures on $S$-structures over $D$. Recursive relational Bayesian networks essentially define a mapping of $R$-structures $\mathscr{D}$ into probability measures on $S$-expansions of $\mathscr{D}$. This mapping, however, is only defined for $R$-structures whose interpretations of the symbols in $R$ lead to well-founded recursive definitions of the probabilities for the $r$-atoms ($r \in S$). If, for instance, $R = \{\leq\}$, and $\mathscr{D}$ is an $R$-structure in which there exist two elements $d_1, d_2$, s.t. neither $d_1 \leq d_2$, nor $d_2 \leq d_1$, then (11) does not define a probability measure on $\{r\}$-expansions of $\mathscr{D}$, because the probability of $r(d_1, d_2)$ gets defined in terms of $r(d_2, d_1)$, and vice versa.

As in section 3, for every $r' \in Pa(r) \cup \{r\}$ a formula $pa_{rr'}(\boldsymbol{x}, \boldsymbol{y})$ can be computed that defines for an $R$-structure $\mathscr{D}$ and $\boldsymbol{d} \subseteq D$ the tuples $\boldsymbol{d'} \subseteq D$, s.t. $F_r(\boldsymbol{d})$ depends on $r'(\boldsymbol{d'})$. While in section 3 existential formulas over the empty vocabulary were obtained, for recursive relational networks the $pa_{rr'}$ are existential formulas over $R$.

The definitions of the probabilities $F_r(\boldsymbol{d})$ are well-founded for $\boldsymbol{d} \subseteq D$ iff the relation $\mathscr{D}(pa_{rr}) := \{(\boldsymbol{d}, \boldsymbol{d'}) \mid pa_{rr}(\boldsymbol{d}, \boldsymbol{d'})$ holds in $\mathscr{D}\}$ is acyclic. A recursive relational Bayesian network $N$ thus defines a probability measure on $S$-expansions of those $R$-structures $\mathscr{D}$, for which the relation $\mathscr{D}(pa_{rr})$ is acyclic for all $r \in S$.

The discussion of inference procedures for relational Bayesian networks in section 3 applies with few modifications to recursive networks as well. Again, we can construct an auxiliary network with nodes for ground atoms, using formulas $pa_{rr'}$ and $pa^*_{rr'}$. The complexity of this construction, however, increases on two accounts: first, the existential quantifications in the $pa_{rr'}, pa^*_{rr'}$ can no longer be reduced to mere cardinality constraints. Therefore, the complexity of deciding whether $pa^{(*)}_{rr'}(\boldsymbol{d}, \boldsymbol{d'})$ holds for given $\boldsymbol{d}, \boldsymbol{d'} \subseteq D$ is no longer guaranteed to be independent of the size of the domain $D$. Second, to obtain the formulas $pa^*_{rr'}$ we may have to build much larger disjunctions: it is no longer sufficient to take the disjunction over all possible paths from $r'$ to $r$ in the network structure of $N$. In addition, for every relation $\bar{r}$ on these paths, the disjunction over all possible paths within $\mathscr{D}(pa_{\bar{r}\bar{r}})$ has to be taken. This amounts to determining the length $l$ of the longest path in $\mathscr{D}(pa_{\bar{r}\bar{r}})$, and then taking the disjunction over all formulas $pa^i_{\bar{r}\bar{r}}(\boldsymbol{x}, \boldsymbol{y}) := \exists \boldsymbol{z}_1, \ldots, \boldsymbol{z}_i(pa_{\bar{r}\bar{r}}(\boldsymbol{x}, \boldsymbol{z}_1) \wedge \ldots \wedge pa_{\bar{r}\bar{r}}(\boldsymbol{z}_i, \boldsymbol{y}))$ with $i < l$. As a consequence, the formulas $pa^*_{rr'}$ are no longer independent of the structure $\mathscr{D}$ under consideration.

## 5 CONCLUSION

In this paper we have presented a new approach to deal with rule-like probability statements for nondeterministic relations on the elements of some domain of discourse. Deviating from previous proposals for formalizing such rules with a logic programming style framework, we here have associated with every relation symbol $r$ a single probability formula that directly defines the probability distribution over interpretations of $r$ within a Bayesian network. The resulting framework is both more expressive and semantically more transparent than previous ones. It is more expressive, because it introduces the tools to restrict the instantiations of certain rules to tuples satisfying certain equality constraints, and to specify complex combinations and nestings of combination functions. It is semantically more transparent, because a relational Bayesian network directly defines a unique probability distribution over $S$-structures, whereas the semantics of a probabilistic rule base usually are only implicitly defined through a transformation into an auxiliary Bayesian network.

Inference from relational Bayesian networks by auxiliary network construction is as efficient as inference (by essentially the same method) in rule based formalisms. It may be hoped that in the case where this inference procedure seems unsatisfactory, namely, for large domains most of whose elements are not mentioned in the evidence, our new representation paradigm will lead to more efficient inference techniques.

## Acknowledgments

I have benefited from discussions with Daphne Koller who also provided the original motivation for this work. This work was funded in part by DARPA contract DACA76-93-C-0025, under subcontract to Information Extraction and Transport, Inc.

## References

Breese, J. S. (1992), "Construction of belief and decision networks", *Computational Intelligence*.

Glesner, S. & Koller, D. (1995), Constructing flexible dynamic belief networks from first-order probabilistic knowledge bases, *in* "Proceedings of ECSQARU", Lecture Notes in Artificial Intelligence, Springer Verlag.

Haddawy, P. (1994), Generating bayesian networks from probability logic knowledge bases, *in* "Proceedings of the Tenth Conference on Uncertainty in Artificial Intelligence".

Halpern, J. (1990), "An analysis of first-order logics of probability", *Artificial Intelligence* **46**, 311–350.

Koller, D. & Halpern, J. Y. (1996), Irrelevance and conditioning in first-order probabilistic logic, *in* "Proceedins of the 13th National Conference on Artificial Intelligence (AAAI)", pp. 569–576.

Ngo, L., Haddawy, P. & Helwig, J. (1995), A theoretical framework for context-sensitive temporal probability model construction with application to plan projection, *in* "Proceedings of the Eleventh Conference on Uncertainty in Artificial Intelligence", pp. 419–426.

Poole, D. (1993), "Probabilistic horn abduction and bayesian networks", *Artificial Intelligence* **64**, 81–129.